\documentclass[11pt]{article}
\usepackage[margin=1in]{geometry}
\usepackage[T1]{fontenc}
\usepackage[utf8]{inputenc}
\usepackage{lmodern}
\usepackage{microtype}
\usepackage{amsmath,amssymb}
\usepackage{booktabs}
\usepackage{array}
\usepackage{tabularx}
\PassOptionsToPackage{hyphens}{url}
\usepackage[hidelinks]{hyperref}
\newcolumntype{Y}{>{\raggedright\arraybackslash}X}
\title{Delivery, Not Storage: Cue-Anchored Working Memory\\as a Harness Property for Coding Agents}
\author{Swapnanil Saha\\Independent\\\texttt{swapnanilsaha26@gmail.com}}
\date{July 2026}

\begin{document}
\maketitle

\begin{abstract}
Coding agents ship with one kind of memory: documents. Instruction files,
plan artifacts, and auto-written memory directories are all \emph{deliberately
authored, deliberately retrieved}: the agent must choose to write them and
choose to read them back. Human expertise runs on a second tier that never
gets written down: situationally-bound operational facts (gotchas,
locations, ``how things are done here'') encoded as a side effect of doing
the work and retrieved involuntarily when the situation cues them. We argue
this second tier is the load-bearing one for long-running agents, and that
it must be a \textbf{harness property, not an agent choice}.

We make four contributions. (1) A two-tier design theory grounded in the
cognitive literature on memory offloading, incidental encoding, and
event-based prospective memory, with each mechanism mapped to an
architectural requirement. (2) A cue-anchored memory model: memories carry
first-class trigger conditions over a composable vocabulary \{path, symbol,
semantic, event, temporal\}, evaluated deterministically by the harness at
the moments that structure an agent's epistemic life, a composition no
surveyed academic or shipped system provides. (3) A controlled evaluation
on a naturalistic coding task showing that voluntary memory use is $\sim$zero
even when the store is pre-seeded with task-relevant knowledge (0 memory
operations in 114 turns), that deterministic injection delivered in every
injection-equipped seeded run ($n=3$) with zero false-alarm fires across audit-logged trigger
evaluations, and that 39\% of intra-session re-reads re-retrieve content
the session had already paid for before a compaction boundary. (4) A
repeated-compaction decay probe: ten facts held only in conversation
vanish at the first summary and stay absent from 106 of 108 ($k=108$ forced
compactions); by the closing phase the deprived agent is grepping the
harness's own session files off disk to reconstruct them --- hand-building,
against standing instruction, exactly the tier the harness never gave it.
The same ten facts, injected from a harness-owned store, arrive intact
through $k=138$ --- 139 audit-logged deliveries: the launch plus every
single compact-resume (138/138) --- while the final
summary carries zero of them: the agent's memory never depended on what
the summarizer chose to keep. Delivery, not storage, is the product: the
reliable memory channel for agents is the one the agent never has to
think about.
\end{abstract}

\section{Introduction}
\label{sec:intro}

A competent engineer who joins a project reads the wiki once, then stops
needing it. What makes them effective six months in is not the wiki --- it
is the accumulated mass of small operational facts that never got written
down: this test is flaky under that flag; releases go through CI only; the
catch-all route must stay last in that file. Nobody decides to memorize
these. They are encoded as a side effect of doing the work, and they
return uninvited, exactly when the situation cues them.

Current coding agents have the wiki and nothing else. The instruction file
(CLAUDE.md and its siblings), the plan artifact, the auto-written memory
directory --- every persistent surface an agent has is a \emph{document}:
deliberately written, deliberately retrieved. The second tier does not
exist. Each session is a competent worker with total amnesia who can read
the wiki.

The obvious fix (give the agent memory tools) has been tried, and it
measurably fails. MEMTRACK~\cite{memtrack} benchmarks memory-equipped agents and
documents systematic failure to utilize memory across long horizons.
TriggerBench~\cite{triggerbench} shows the deeper reason: acting on a latent
standing constraint when its trigger appears (prospective memory) is much
harder for models than retrieving stated facts (retrospective memory), and
prospective performance \emph{decays with context length}, precisely the
regime long-running agents occupy. Our own controlled runs replicate this
naturalistically and extend it to the strongest case: an agent whose store
was \textbf{pre-seeded with facts directly relevant to its task}, with
connected tools and explicit guidance, made zero memory calls in 114 turns
(\S\ref{sec:adoption}). Voluntary memory does not happen. Vendors concede this in their
own products: Anthropic's memory tool force-injects an all-caps directive
to check memory before anything else; ChatGPT's memory does automatic
capture and automatic injection, with no voluntary lookup tool at all.

The distinction that organizes this paper is between memory as \textbf{content}
and memory as \textbf{control}. A memory system operating in the \emph{content
plane} adds material to the context (files, instructions, protocols)
and relies on the model's compliance to turn that material into memory. A
system operating in the \emph{control plane} participates in constructing the
context itself: it decides, deterministically, what enters the window and
when. Field data makes the stakes concrete: a heavy user of an agentic
coding CLI, documenting 59 context compactions over 26 days, built a
three-tier file vault with standing orders and a compaction watcher~\cite{fieldreport}:
an entire memory architecture in the content plane, operable only
through relentless manual enforcement, because the control plane offered
nothing. Everything they hand-built traces the shape of a missing harness
feature.

This paper describes that feature and measures what its absence costs.
Contributions:

\begin{itemize}
\item \textbf{C1 --- Design theory (\S\ref{sec:twotiers}).} A two-tier account of engineering memory
  grounded in the cognitive literature, in which each mechanism
  (incidental encoding, cue-driven retrieval, endorsement, resistance to
  documentation) maps to a falsifiable architectural requirement.
\item \textbf{C2 --- Mechanism (\S\ref{sec:model}).} A cue-anchored memory model: per-memory
  standing trigger conditions over \{path, symbol, semantic, event,
  temporal\}, evaluated deterministically harness-side, with budgeted,
  provenance-framed, staleness-checked delivery. No surveyed system
  composes these properties (\S\ref{sec:related}).
\item \textbf{C3 --- Controlled evaluation (\S\ref{sec:correctness}--\ref{sec:singleboundary}).} Twelve graded runs of a real
  feature task under asserted launch postures: voluntary use is $\sim$zero
  even seeded (0 ops in 114 turns); deterministic delivery works through
  two independent channels with zero false alarms; count-based
  efficiency signals are consistently positive and honestly labeled
  directional; 39\% of re-reads are cross-compaction re-purchases.
\item \textbf{C4 --- Decay probe (\S\ref{sec:decay}).} A forced-compaction experiment ($k=108$ and
  $k=138$) isolating what repeated summarization does to conversation-held
  facts, with token-level survival curves, provenance forensics on every
  endpoint, and two second-order findings: continuation summaries
  propagate confabulated task state, and a harness tuned for memoryless
  operation actively fights a memory-carrying configuration.
\end{itemize}

\S\ref{sec:twotiers} grounds the design; \S\ref{sec:model} specifies the mechanism; \S\ref{sec:surface} describes the
implementation surface; \S\ref{sec:eval} reports the evaluation; \S\ref{sec:related} positions against
related work; \S\ref{sec:threats} analyzes threats to validity; \S\ref{sec:conclusion} concludes.

\section{Two tiers, and why the second must be harness-owned}
\label{sec:twotiers}

\begin{table}[h]
\centering
\small
\begin{tabularx}{\textwidth}{@{}p{2.2cm}YY@{}}
\toprule
 & \textbf{Tier 1 --- documents} & \textbf{Tier 2 --- brain memory} \\
\midrule
Content & Decontextualized, durable, shareable: specs, ADRs, runbooks & Situationally-bound operational facts: gotchas, locations, tacit fluency \\
Encoding & Deliberate act of authorship & Automatic side effect of the work \\
Retrieval & Deliberate lookup & Cue-driven, largely involuntary \\
Failure mode & Staleness; write-only archives & Lost when the context window resets \\
\bottomrule
\end{tabularx}
\end{table}

\emph{Brain memory} is the informal name we use for the second tier; the
mechanism this paper specifies and evaluates for it is cue-anchored
working memory (\S\ref{sec:model}).

The mapping from cognitive mechanism to design requirement is generative,
not decorative; each row forced an architectural decision:

\begin{itemize}
\item \textbf{Encoding is incidental.} In levels-of-processing and
  incidental-learning studies, subjects given no instruction to remember
  recalled as well as deliberate memorizers~\cite{craik,hyde}. (Deliberate
  generation does aid retention when it happens~\cite{slamecka}; the design
  premise is only that it cannot be relied on to happen.) Humans don't
  call \texttt{remember()}. A design that depends on the model volunteering a
  save fights the grain of the only system known to do memory well.
  $\rightarrow$ Capture must not depend on agent initiative (the open half; \S\ref{sec:threats}).
\item \textbf{Retrieval is cue-driven and mostly involuntary.} Encoding
  specificity~\cite{tulvingthomson}; involuntary autobiographical memories outnumber
  voluntary recall roughly 2:1 in daily-life sampling~\cite{berntsen};
  event-based prospective memory is more automatic than time-based~\cite{mcdaniel}.
  $\rightarrow$ The default delivery channel is injection at the cued moment; voluntary
  lookup is the fallback, never the load-bearing path.
\item \textbf{An external store counts as memory only under the extended-mind
  criteria}~\cite{clark}: reliably available, near-zero effort, \emph{automatically
  endorsed}. Clark argued web search fails the endorsement criterion:
  results get evaluated, not trusted. The offloading literature
  measures the other half of the bargain: externalizing reliably frees
  capacity and reshapes what stays internal (people retain \emph{where}
  rather than \emph{what}~\cite{risko,sparrow}), and saving one item measurably improves
  memory for the next~\cite{storm}; its group-level analog is transactive
  memory~\cite{wegner}. Injection makes a store function as memory; lookup makes
  it function as a reference document. Same content, different
  delivery, different cognitive role.
\item \textbf{What resists being written down.} The episodic/semantic and
  declarative/procedural splits~\cite{tulving,squire} explain why writing a
  situationally-bound fact strips exactly the cues that make it
  retrievable. ``This config is load-bearing'' is useful only when you are
  staring at that config --- which is why humans never documented it, why
  a flat memory file serves it badly, and why a \emph{cue-anchored} record
  serves it well.
\end{itemize}

The content/control distinction of \S\ref{sec:intro} is the systems restatement of the
endorsement criterion: only control-plane delivery gives a store the
trust posture and the reliability of memory. Content-plane systems,
however elaborate, inherit the failure rate of model compliance, which
\S\ref{sec:eval} measures.

\section{The cue-anchored memory model}
\label{sec:model}

A memory is a tuple \textbf{(content, kind, triggers, scope, decay)}.

\textbf{Kinds} map to the cognitive taxonomy: \texttt{directive} (standing
prospective intention that must never miss), \texttt{gotcha} (cue-anchored
caveat), \texttt{finding} (semantic fact), \texttt{task} (working state), \texttt{reference}
(pointer).

\textbf{Triggers} are per-memory standing conditions over a composable
vocabulary: \textbf{path} (glob over touched files), \textbf{symbol} (a named code
entity is referenced; requires a code symbol graph), \textbf{semantic}
(similarity of current activity above a floor), \textbf{event} (session
start, prompt submit, pre-edit, pre-run, pre-compaction,
post-compaction), \textbf{temporal} (not-before, cooldown). Conjunction
within a trigger, disjunction across triggers. Kinds define default
delivery when no explicit trigger is present: directives fire at session
start and post-compaction; high-priority task notes at session start
(including every compact-resume); findings enter prompt-time semantic
recall over a similarity floor.

\textbf{Evaluation is deterministic and harness-side.} No model judgment sits
in the delivery path --- the failure TriggerBench documents (models cannot
self-trigger) and the fallibility Devin's own documentation concedes for
semantically-matched knowledge items are both designed out. Determinism
also buys auditability: every evaluation, fire, and suppression is a
logged system event (\S\ref{sec:mechanism} gates its mechanism claims on this audit log),
which no model-mediated selection can offer.

\textbf{Delivery discipline.} Injection is budgeted (two-tier: a compact
index by default, full content on demand or for cue-anchored fires); a
per-session fire ledger dedups repeats, and the ledger resets at
compaction boundaries so anchored facts re-arm in the new window;
injected content carries provenance framing (``recorded by an AI session,
not human-endorsed --- verify'') so memory stays the epistemic surface and
instructions stay the imperative one; staleness is checked against
ground truth, so if the anchored file changed after the note was written
the injection carries an explicit warning. Stale-memory poisoning,
injection bloat, and false-alarm rate are first-class design constraints
(the poisoning literature demonstrates adversarial cue-anchored
firing, planted items persisting dormant until a later trigger, and
finds lightweight rule- and filter-based defenses insufficient~\cite{sleeper}).

\section{Implementation surface}
\label{sec:surface}

The model is implemented in Vectr, an open-source local-first
code-indexing daemon with a working-memory store (see Artifact
Availability). Delivery integrates with the agent harness (Claude
Code, an agentic coding CLI) at two alternative points:

\begin{enumerate}
\item \textbf{Native hooks.} The harness's lifecycle hooks (session start, user
   prompt submit, pre-tool-use, pre-compaction) invoke a hook binary
   that consults the daemon and returns context to inject. This is the
   channel a first-class ``memory mode'' would own.
\item \textbf{API proxy.} A transparent proxy on the model-API path injects
   context deterministically into requests. Model- and product-agnostic;
   used here as the mechanism-validation control.
\end{enumerate}

Both channels deliver the same store. The evaluation's design point: if
the same knowledge, delivered through two unrelated mechanisms, produces
the same behavioral signature --- and its absence through a third
(voluntary tools) produces none --- the effect is the delivery
architecture, not the implementation.

\section{Evaluation}
\label{sec:eval}

\textbf{Setup.} One naturalistic task on a pinned Apache Camel checkout
($\sim$169k indexed chunks): implement the \texttt{reverse} option for the
stream-mode Resequencer EIP (Enterprise Integration Pattern), a real
feature requiring cross-module exploration (model layer, processor
engine, reifier wiring, catalog/schema surfaces), gated by a
harness-provided acceptance test that must pass unmodified plus the
resequencer-scoped regression set (\texttt{-Dtest='*Resequenc*Test'}, 42
tests). All arms run the same agent product, same model (Claude
Sonnet 5, \texttt{claude-sonnet-5}), same prompt, same corpus SHA. Canonical
wall time is the transcript's \texttt{duration\_ms} (sleep-immune). Every
run's MCP (Model Context Protocol) posture, guidance surface, memory
starting state, and gate-test fingerprint are asserted at launch and
archived. Seeded arms start from four exploration-grade notes: facts a
prior exploration session would plausibly have stored (locations,
semantics, the guard site), never solution steps; one note is a gotcha
anchored to the file containing the guard.

Arms (9 pilot runs + 3 native-channel runs, all graded):

\begin{table}[h]
\centering
\small
\begin{tabularx}{\textwidth}{@{}llYY@{}}
\toprule
\textbf{arm} & \textbf{n} & \textbf{memory surface} & \textbf{delivery} \\
\midrule
A & 3 & none & --- \\
B & 2 & store (cold) + tools + guidance & voluntary \\
C & 3 & store + tools + proxy, cold store & injection (nothing to inject) \\
CS & 1 & store seeded & proxy injection \\
H & 2 & store seeded & \textbf{native hooks} (session-start / prompt-submit / pre-tool-use / pre-compact) \\
V & 1 & store seeded & voluntary (hooks stripped; control for H) \\
\bottomrule
\end{tabularx}
\end{table}

The repeated-compaction decay probe (\S\ref{sec:decay}) uses a separate workload on
the same corpus and defines its own arms (N, M) there.

\subsection{Correctness: no harm}
\label{sec:correctness}

12/12 graded runs pass the acceptance test (byte-unmodified,
fingerprinted) and the regression suite. Injection, proxy or native,
never cost correctness. One archive literalism a reader will hit
first: the per-run \texttt{sha-check} line reads FAILED in every run
directory, because Camel's build formatter re-wrapped two javadoc
comment lines of the gate test in every run; per-run transcript audits
confirm zero agent edits to the test, and the canonical test was
restored byte-for-byte before the grading run (documented in the
CS/H/V grade files; the pilot runs are evidenced by per-run diff
scopes). (Six further matrix launches did not gate, all retained with
diagnosis: three preflight aborts on two harness-surfaced product bugs
(one store-clear defect, two on a REST-starvation availability bug),
one hook-probe preflight abort, one operator pause, and one pilot
invalidated at the launch-posture assert.)

\subsection{Adoption: voluntary memory does not happen}
\label{sec:adoption}

\begin{itemize}
\item Unseeded equipped runs (B, C): voluntary memory writes 0--1 across all
  five runs, despite 32 memory-guidance mentions in the workspace
  instruction files and a verified connected tool surface. The native auto-memory
  directory was never created in any run.
\item \textbf{The seeded-voluntary control (V) is the strongest form of the
  finding}: same four task-relevant notes in the store as the H arms,
  same connected tools, same guidance --- the agent made \textbf{zero memory
  calls in 114 turns}. Knowledge present in a store contributes nothing
  by itself.
\item Injection begets engagement: the seeded-proxy run (CS) produced the
  matrix's only memory-hygiene loop (recall by id, \emph{forget} of the
  gotcha its implementation had just obsoleted, then \emph{remember} of a
  completion note). The loop came after its first two voluntary calls
  timed out on the availability bug of \S\ref{sec:correctness}. The native runs reproduced the pattern (H1: expansion
  of all three injected index entries by id within seconds of injection,
  later a semantic recall returning staleness-flagged notes, then
  supersession of the obsolete gotcha; H2: one immediate semantic
  recall $\sim$17 s in, then working from the injected content without
  expanding the index by id). Voluntary memory operations occur \emph{after
  and because of} delivery, not instead of it.
\end{itemize}

\subsection{Mechanism: deterministic delivery, proven on two channels}
\label{sec:mechanism}

\textbf{Proxy (CS):} 262 API requests: 241 reached the injection decision ---
5 injected, 236 correctly skipped --- and 21 (by subtraction; the
status file carries no per-type counter) did not; zero injection or
upstream errors. The proxy's own status counters are the published
ledger for this channel (\path{proxy-status.json}); the grading-time
daemon audit additionally recorded six \texttt{PROACTIVE\_INJECT} events
anchoring the seeded notes, quoted in the published grade file --- one
more than the proxy's injected count, plausibly the pre-launch
injection probe, now unreconcilable because that daemon log predates
this archive's excerpt policy and was not retained.

\textbf{Native hooks (H$\times$2):} the prompt-submit hook injected a three-note
index at launch in both runs, behaviorally confirmed in H1 by the
agent's first three tool calls being recalls of exactly the three
injected note ids (knowable from nowhere else). The cue-anchored gotcha
\textbf{fired inside both runs on the agent's first touch of the anchored
file} ($\sim$20 s and $\sim$7 min in), per the daemon audit log; subsequent
touches were suppressed by the per-session fire ledger, and the ledger
reset at each compaction boundary, as designed. False-alarm
injections: zero in both runs, across 40 and 35 audit-logged trigger
evaluations respectively (the only other fire in each window is the
harness's own pre-launch hook probe). One delivery honesty note: the
injected index carried three of the four seeded notes. The index
budget dropped one (a known ranking limitation logged as product
work); the agent recovered that note's content from code exploration.

A methodological note for replication: the agent transcript records
session-start hook payloads but not prompt-submit or pre-tool-use
payloads; for those, the daemon audit log is the evidentiary surface
for injection claims (sanitized per-run excerpts are published as
\path{daemon-audit-excerpt.log}).

\subsection{Value, honestly priced}
\label{sec:value}

Count-based metrics separate the arms more cleanly than wall-clock,
which carries API-latency and agent-path noise:

\begin{table}[h]
\centering
\small
\begin{tabularx}{\textwidth}{@{}l>{\raggedleft\arraybackslash}X>{\raggedleft\arraybackslash}Xr@{}}
\toprule
\textbf{metric} & \textbf{vanilla mean ($n=3$)} & \textbf{tool-equipped mean ($n=6$, pilot; store cold in 5 of 6)} & \textbf{$\Delta$} \\
\midrule
total tool calls & 137.3 & 79.8 & $-$42\% \\
grep/find calls & 48.3 & 22.2 & $-$54\% \\
cost & \$7.20 & \$5.08 & $-$30\% \\
turns & 106.3 & 77.0 & $-$28\% \\
wall (s) & 2123.6 & 1432.8 & $-$33\% \\
\bottomrule
\end{tabularx}
\end{table}

Five of the six pooled runs ran with an empty store, so this table
prices the platform (semantic-search tools plus guidance), not memory
delivery; it bounds the tool-surface effect and nothing in this paper's
memory argument rests on it. The memory measurement is the
knowledge-held-constant comparison (same seeds):

\begin{table}[h]
\centering
\small
\begin{tabular}{@{}llrrr@{}}
\toprule
\textbf{run} & \textbf{delivery} & \textbf{turns} & \textbf{cost} & \textbf{wall (s)} \\
\midrule
H1 & hooks & 95 & \$5.22 & 1454 \\
H2 & hooks & 102 & \$6.09 & 1187 \\
V1 & voluntary & 114 & \$6.23 & 1297 \\
\bottomrule
\end{tabular}
\end{table}

Both H runs finished in fewer turns and cheaper than V; the native
channel also beat the pilot's proxy run (CS: 106 turns, \$6.57, 1629 s)
on every count. At $n=2$ vs $n=1$ (and $n=3$ vs $n=6$ in the pilot), with
within-arm wall spreads of 2.4--4.4$\times$, none of this clears a significance
bar, and we make no headline speed claim. The direction is consistent
across every metric and both channels; a defensible percentage would
need $\sim$15--20 runs per arm.

\subsection{Re-exploration waste}
\label{sec:waste}

Over the nine pilot transcripts (analyzer and per-run data published;
cf.~\cite{sweexplore} on SWE-agent exploration behavior): \textbf{61 intra-session
re-reads}, of which \textbf{24 (39\%) re-read content the session had
already read before a compaction boundary}; $\sim$78k result tokens
re-paid (chars/4 proxy); pooled re-exploration share 10.9\% of
exploration calls (range 0--29.1\%). The distribution is the finding:
the three heaviest runs hold 55 of 61 re-reads and 21 of 24
cross-compaction re-reads (the remaining three sit in a fourth
compacted run); every pilot run compacted at least once, yet four
finished with zero re-reads --- the boundary is where waste concentrates
when it occurs, not a guarantee that it does. The worst single run re-paid $\sim$31.6k
tokens (a sixth of a context window) re-retrieving content it had
already paid for. Waste concentrates exactly at context-loss
boundaries, which is what deterministic post-compaction re-injection
targets.

\subsection{Compaction, single boundary}
\label{sec:singleboundary}

All twelve graded runs auto-compacted at least once mid-flight; all
finished green. A single compaction is near-lossless for task continuity:
the harness's continuation summary carries it, with or without a memory
tier. What the memory tier measurably added at the boundary:
post-compaction recall returned the seeded notes with live staleness
warnings (the agent had since edited the anchored file; a frozen
summary cannot self-invalidate), the agent superseded the obsolete
note, and the trigger-ledger reset re-armed cue-anchored delivery for
the new window.

Under \emph{repeated} compaction --- the one-conversation-per-project usage
pattern (field report: 59 compactions in 26 days~\cite{fieldreport}) --- in-context
facts must survive re-selection at every pass, endure lossy
extraction~\cite{proactivemem,recmem} and iterative re-encoding drift~\cite{ssgm,recursummary}, and compete
for a fixed summary budget against a growing history. An externalized store is immune by
construction. \S\ref{sec:decay} measures both halves.

\subsection{Repeated compaction: the decay probe}
\label{sec:decay}

\textbf{Workload.} A second harness drives the same pinned corpus through a
load sized to compact continuously: one seeding phase, sixteen
exhaustive file-audit phases over a 64-file corpus, and a closing
phase, executed by a small fast model (Claude Haiku 4.5) as one
resumed session with the auto-compaction window pinned at 100k tokens
(the CLI's floor).

\textbf{Facts.} Ten synthetic operational facts (a build-freeze code, an
incident bridge, a rollback deadline, \ldots) are presented exactly once,
in phase 0, in a fixture file that is \textbf{deleted immediately
afterward}: removing the only re-readable source defeats the
harness's post-compaction file restoration, so any later appearance of
a fact must have survived in conversation or in a memory tier. Every
phase prompt restates the tool constraints (file access through the
Read tool only --- no shell; no subagents); the closing phase requires
writing the ten facts into the audit report.

\textbf{Grading.} Token-level: presence of each fact token in every
continuation summary (the survival curve) and in the final report
(endpoint), plus provenance forensics over the full session transcript
and the daemon audit ledger.

\textbf{Arms.} \textbf{N} runs with no memory tier. \textbf{M} runs the identical
workload with the ten facts seeded as one-line notes in the
working-memory store, re-injected by the session-start / prompt-submit
hooks at every compact-resume.

\begin{table}[h]
\centering
\small
\begin{tabularx}{\textwidth}{@{}p{5.4cm}YY@{}}
\toprule
 & \textbf{arm N (no memory tier)} & \textbf{arm M (injected store)} \\
\midrule
compactions $k$ & 108 & 138 \\
turns / cost & 1,235 / \$21.26 & 1,489 / \$28.94 \\
summaries carrying 10/10 facts & 2 (after an accidental re-surfacing) & 53 \\
summaries carrying 0/10 facts & 106 & 85 \\
summaries carrying a strict subset & 0 & 0 \\
summary 1 (first boundary) & \textbf{0/10} & \textbf{0/10} \\
final summary & 0/10 & \textbf{0/10} \\
endpoint (10 facts in report) & 10/10 --- \textbf{invalid} (transcript archaeology) & 10/10 --- \textbf{clean} (injection-sourced) \\
fact-bearing deliveries (daemon audit ledger) & 0 & 139 (all ten facts each; 157 evaluations logged) \\
\bottomrule
\end{tabularx}
\end{table}

\textbf{Figure 1 --- survival regimes.} Every summary is all-or-nothing
(0/10 or 10/10 fact tokens; no strict subset ever occurred in either
arm):

\begin{table}[h]
\centering
\small
\begin{tabularx}{\textwidth}{@{}llY@{}}
\toprule
\textbf{arm} & \textbf{summaries} & \textbf{regime} \\
\midrule
N & 1--29 & 0/10 \\
N & 30--31 & 10/10 --- an orientation grep accidentally re-surfaced the block \\
N & 32--108 & 0/10 \\
M & 1--3 & 0/10 --- dropped despite injection into the summarized context \\
M & 4--66 & 10/10 in 53 of 63 \\
M & 67--138 & 0/10 --- 72 consecutive, while deliveries continued on schedule \\
\bottomrule
\end{tabularx}
\end{table}

\textbf{The summary channel drops everything at the first boundary.}
In both arms, summary 1 carried 0/10 fact tokens; in N, 106 of 108
summaries carried 0/10.
The two exceptions measure re-injection half-life: mid-run, one of the
agent's own orientation greps accidentally re-surfaced the fact block
into context; the next two summaries carried 10/10; one
re-summarization pass later the facts were gone again. A fact
re-entering context survives approximately one further pass. Selection
is also all-or-nothing. Across both arms, no summary ever carried a
strict subset of the ten facts: the summarizer either folds the block
in or drops it whole.

\textbf{Denied a memory tier, the agent builds one (N endpoint).} Arm N's
final score was 10/10 --- and provenance forensics invalidate it. In the
closing phase the agent ran a deliberate archaeology chain over the
harness's own persistence: grepping its session JSONL and the
harness's persisted tool-result files for the fact markers, then
transcribing the hits: 260 shell commands in total (38 fact
archaeology against the harness's persisted session artifacts, 152
corpus/orientation, 70 report operations), against a ban restated in
every phase prompt. (First shell use appears only after 15
summaries, so the curve's early region is a clean pure-summary-channel
measurement.) The asymmetry the agent exploited is instructive: the
harness restores \emph{files} after compaction and re-reads root
instruction files, but restores nothing the conversation knew. Given
no sanctioned memory tier, the model hand-built an episodic one from
the harness's disk artifacts. The endpoint tie with M is an artifact
of this prompted closing phase --- it measures what each arm could
produce when explicitly asked. The operative quantity for real work is
mid-run availability, where the arms are not tied: N's post-compaction
window carried 0/10 fact tokens at 106 of 108 boundaries, while M's
received all ten by injection at every one of its 138.

\textbf{The injected store survives everything (M).} Final score 10/10 with
clean provenance: zero fact-targeted shell commands, an empty
disk-memory directory, no early fact write into the report, and,
decisively, a final continuation summary carrying 0/10: had the
endpoint depended on the summary channel, it would have scored zero.
The daemon's audit ledger records 157 session-start evaluations, of
which 139 delivered all ten facts into the run: one at every single
compact-resume (138/138) plus the first phase launch. (The ledger's
one other fact-bearing line, 1.5 s before the launch's, is the
runner's pre-launch hook probe, payload archived as
\path{hook-probe-sessionstart.json}, never part of the session.) The 17
remaining mid-session phase-launch evaluations fired zero notes (16)
or one non-seed note (1), the per-session fire ledger of \S\ref{sec:model}
suppressing redundant re-injection while the facts were still
in-window, and the post-kill relaunch logged no evaluation at all. Delivery fired exactly when a
fresh window needed it and nowhere else. Each delivery ran $\sim$240
tokens (mean 969 chars of injected text, chars/4 as in \S\ref{sec:waste}), $\sim$34k
injected tokens across the run; all 139 are visible in the session file as
hook-attachment pairs carrying 10/10 fact tokens. The tier's price,
measured: the delivery overhead, the guard-forced tighter read cap
(below), and the late-run confabulation spiral put M at +21\% turns
and +36\% cost versus N. One disclosure: of
M's 321 shell results, exactly one (an orientation grep over its own
session file for phase-10 task state) incidentally echoed fact tokens
back into context; it sits over sixty compaction boundaries before the
endpoint and cannot reach it. Every fact-bearing carrier in the
endpoint window is a hook attachment.

The survival curve is bimodal with a regime change (Figure 1): even
with the facts injected into the very context being summarized, the
first three summaries carried none; summaries 4--66 folded them in 53
times out of 63; then the entire back half of the run (72 consecutive
summaries) carried none, while deliveries continued to arrive on
schedule. We did not predict the regime change and do not fully
explain it (plausibly the growing report and the confabulation spiral,
below, changed what the summarizer selected); either reading supports
the same conclusion: summarizer selection is unreliable \emph{independently
of fact availability} --- and with a memory tier, nothing depends on it.
The injection-begets-engagement pattern of \S\ref{sec:adoption} also replicates under
decay: the M agent voluntarily issued 11 remember calls (9 landed in
the audit ledger) and 9 recalls mid-run, against zero voluntary
operations in the seeded-voluntary control. (The prompt-submit index tier separately delivered partial
fact coverage, 3/10 tokens early and 1/10 late --- the budgeted-index
lesson of this section's closing note, caught live.)

\textbf{What nobody seeded still decayed (``Phase 52: Files 203--206'').}
Only the ten facts were seeded. The task state (which files belong to
which phase) lived in conversation, and under 138 compactions it
decayed into confabulation: the agent skipped two entire audit phases,
invented an arithmetically self-consistent phase numbering deep into
the run (``Phase 52: Files 203--206'' --- in a sixteen-phase, 64-file
audit), re-audited completed files under the invented scheme, and
spent 37 transcript-archaeology greps, nine memory recalls, and three
(forbidden) subagent launches hunting a task specification that never
existed. The hunt even left the corpus: the agent twice \texttt{Read} the
memory system's own design document off the operator's disk, reasoning
it ``might have the canonical specification'' --- a real path whose first
appearance in the recorded stream is the agent's own thinking, with no
recorded source (the file contains no fact tokens; both reads appear
in the audit ledger). The continuation summaries did not correct the false task
model --- they propagated it across every boundary. (The pilot showed
the same signature at $k=2$: a first summary asserting ``Memorization:
Maintained all 10 DECAY-FACTS'' while carrying zero of them.) Summaries
consolidate the model's current beliefs; they do not validate them.
Audit completeness (declaration recall against a regex ground truth
over the 64 source files) came out 0.70 for M vs 0.93 for N, with
the confound that M ran under a tighter read cap, forced on it by the
harness (next paragraph). Both scores are from the format-robust
regrade published per-run as \texttt{decay-grade-v4-regrade.txt}.

(Archive note: the original grade files retain three at-a-glance
flags a reader will hit first, each decoded in the run log. The
recall column's 0.00 was a section-binding bug in the original grader
(fixed; N's score is unchanged by the fix), ``SAMPLING/FABRICATION
SUSPECT'' is the confabulated-numbering finding itself, and ``RUN
INVALID for the in-context-only premise'' is the no-memory-arm
contamination rule firing, correctly by its own logic, on the
memory arm's sanctioned injections.)

\textbf{The harness fights the memory-carrying configuration.} The agent
product's autocompact thrash guard aborts a call when context refills
within three turns of a compact three times consecutively. The guard
is not memory-specific: it also killed arm N once mid-run (phase 6,
after 49 boundaries; relaunched and completed, still at the 8k read
cap). The difference is structural margin. The memory-equipped arm
carries fixed per-cycle context the naive arm does not: MCP tool
schemas re-loaded per resumed call, guidance files, per-prompt
injections --- a fixed context overhead the guard reads as refill
pressure. At the arms' shared 8k cap the guard killed M's first audit
phase fourteen turns in and the launch was scrapped; running required
lowering only M's read cap to 6k tokens, and even at 6k the guard
killed the final audit phase as the growing report tightened the
squeeze. One kill for N across eighteen phases at 8k; for M, a
scrapped launch at 8k and one kill even at 6k. A harness tuned for
memoryless operation treats the context profile of a memory-carrying
configuration as elevated thrashing risk --- the operational form of
this paper's content/control argument, measured.

A delivery-design lesson from the probe's preflight: injection
surfaces that ship an index tier (title line now, expand on demand)
deliver only the title at session boundaries --- a fact must ride the
index line itself or it does not travel. The probe's seeds are
one-line notes for exactly this reason.

\section{Related work}
\label{sec:related}

A systematic sweep across four lanes --- prospective/trigger-conditioned
memory, memory-as-governance, agent-memory surveys 2024--2026, and
shipped product mechanisms --- narrows the novelty claim to a specific
composition. Per-item triggers exist in fragments. Devin's knowledge
items carry per-item ``trigger descriptions'', semantically matched
and fallible by the vendor's own documentation. Pathrule ships
deterministic hook-time delivery restricted to path globs over
team-curated knowledge. The rules ecosystem (path-scoped instruction
files across major AI IDEs) made deterministic per-item path triggers
industry-standard \emph{for static human-authored configuration};
OpenHands' repository microagents extend it to per-item keyword
triggers evaluated harness-side --- still static, human-authored
knowledge, not agent-written memory. Encoding standing user
preferences as program state~\cite{userascode} is adjacent in spirit (persistent
conditioning of behavior) but again human-scoped. PROJECTMEM's
pre-action gate~\cite{projectmem} is genuine cue-anchored governance,
hardcoded to two derived warning types and evaluated in a
single-developer self-study. General memory architectures (MemGPT~\cite{memgpt},
Mem0~\cite{memzero}, Zep~\cite{zep}, A-MEM~\cite{amem}) offer stores and retrieval
policies, not standing per-item trigger conditions. A recent
systematic study of general-purpose memory systems~\cite{agentnative} maps retrieval
across its subjects as query-time strategy (attention-based, semantic,
topological, agentic routing, hybrid); standing per-item trigger
conditions, evaluated against the agent's ongoing activity, appear
nowhere in its taxonomy.

What no surveyed system has, and this design contributes: (1)
composable multi-vocabulary triggers, \{path, symbol, semantic, event,
temporal\}, as per-memory standing conditions (symbol-conditioned
memory requires a code symbol graph, which no memory system carries);
(2) the trigger-carrying item is \emph{agent-written working memory}, not
static rules or onboarding knowledge; (3) deterministic harness-side
evaluation with an audit surface; (4) the measurements: TriggerBench
and MEMTRACK quantify the motivating failure, not a mechanism; no
shipped or academic system publishes trigger-precision or
injection-utility numbers, and we know of no prior forced-compaction
survival measurement of either the summary channel or an injected
store. The two-tier design theory (\S\ref{sec:twotiers}), used generatively, has no
analog in any surveyed row.

\section{Threats to validity}
\label{sec:threats}

\textbf{Construct.} Fact survival is verbatim token-match; a paraphrased
survival would be missed (none was observed on spot-check, and the
all-or-nothing selection pattern suggests block-level, not token-level,
summarizer behavior). The decay probe's facts are synthetic and carry
an explicit closing-phase obligation --- the probe measures survival and
delivery under compaction pressure, not spontaneous use. Token costs
in \S\ref{sec:waste} use a chars/4 proxy and support relative comparison only.
Audit completeness is measured by declaration recall against a regex
ground truth, which undercounts prose-only audit content. The trigger
vocabulary is exercised unevenly: the graded runs fire path, event,
and semantic triggers; symbol and temporal triggers are implemented
but never fired by any graded run, so C2's composition claim rests on
three of its five elements having measured fires. Capture is the
unevaluated half (\S\ref{sec:twotiers}'s open half): the probe's ten fact notes were
seeded by the harness, and \S\ref{sec:eval}'s voluntary-write counts measure initiative,
not capture quality; capture-side automation is future work.

\textbf{Internal.} Arm M ran at a 6k read cap vs N's 8k, forced by the
thrash guard and reported as a finding, but a confound on the
0.70-vs-0.93 audit-completeness comparison. Tool bans proved
unenforceable in both decay arms (shell archaeology in both; forbidden
subagents additionally in M); we treat constraint non-compliance under
context loss as data, and
gate every endpoint on transcript-level provenance forensics rather
than on compliance. An 11-minute host hibernation occurred mid-run in
arm M (processes restored; session continuous; wall-clock excluded
from claims). The evaluated implementation and the benchmark harness
share an author; mitigations: pinned corpus SHA (\texttt{a543dc64}),
fingerprinted acceptance test, per-run launch-posture assertions
archived, audit-log-gated mechanism claims, all raw transcripts,
graders, and analyzers published.

\textbf{External.} One corpus (Apache Camel), one task family per
experiment, one agent product, one model family; the decay probe uses
a small fast model (Claude Haiku 4.5), chosen to make $k \ge 100$
compactions affordable; decay behavior of larger summarizers may
differ. Camel is public and present in training corpora: if
equivalent resequencer logic exists anywhere upstream, part of the
fix may be pretrained knowledge; this affects all arms equally and
cannot manufacture a between-arm delta, but it shrinks the
exploration burden the arms differ on. The decay probe's pinned 100k
window also halves per-boundary content pressure relative to stock
thresholds; smaller summarized segments should make per-boundary
survival easier, not harder, so the observed first-boundary drop is
conservative on that axis. Results are directional evidence with
asserted controls, not population estimates.

\textbf{Conclusion.} Sample sizes are small throughout (12 graded matrix
runs; one gated run per decay arm, each after protocol repairs
documented in the published run log). We report absolute numbers with
$n$ visible, claim direction rather than effect size, and mark every
percentage in \S\ref{sec:value} accordingly. Contested sources are excluded from
the argument (the diver study's failed replication, longhand-note
effect sizes, unverified vendor claims about re-read token shares).

\section{Conclusion}
\label{sec:conclusion}

The reliable delivery channel for agent memory is the one the agent
never has to think about. Giving models memory tools reproduces the
wiki; giving the harness a cue-anchored tier reproduces the thing
expertise actually runs on. The evaluation is small but its signature
is consistent: voluntary memory use rounds to zero even with a seeded
store; the same store, delivered deterministically at the cued moment
through either of two unrelated mechanisms, arrives every time,
costs nothing in correctness, and begets the only voluntary memory
engagement observed anywhere in the corpus. Under repeated compaction
the case sharpens to a point: the summary channel dropped every
planted fact at the first boundary in both arms --- even the arm whose
context held the facts at summarization time; the agent denied a memory tier
taught itself transcript archaeology to build one from the harness's
disk artifacts; and the injected store delivered all ten facts through
138 compactions while the final summary carried none of them. Memory
for agents is not a content problem. It is a control-plane problem.

\section*{Artifact availability}

All run archives, launch-posture assertions, graders, the
re-exploration analyzer, the decay-probe harness, and per-run
transcripts are published in the Vectr repository
(\url{github.com/swapnanil/vectr}) under \path{research/proactive-gate/}
(matrix runs, decay results \path{decay-N-20260717-030703} and
\path{decay-M-20260717-192249}, protocols, complete run logs including
invalidated launches, per-run daemon audit-log excerpts
\path{daemon-audit-excerpt.log}, and format-robust regrades
\path{decay-grade-v4-regrade.txt}) and \path{research/brain-memory/} (this paper, the
analyzer, and measurement data). Large transcripts are
gzip-compressed; the operator's local paths and one corporate Maven
mirror hostname were rewritten for privacy, with the exact rewrite map
documented in \path{research/README.md}. The daemon implementation ships
from the same repository.

\end{document}